\documentclass[conference]{IEEEtran}  % or journal, draftclsnofoot, onecolumn, etc.

\usepackage{cite}
\usepackage{amsmath,amssymb,amsfonts}
\usepackage{algorithm}
\usepackage{algorithmic}
\usepackage{graphicx}
\usepackage{textcomp}
\usepackage{booktabs}
\usepackage{multirow}
\usepackage{array}
\usepackage{xcolor}
\usepackage{epsfig}
\usepackage{times}

\usepackage{fancyhdr}

\fancypagestyle{firstpage}{%
  \fancyhf{} % clear all header/footer
  \fancyhead[L]{\small Pre-Print: Accepted in MICCAI 2025} % top-left corner
}

\def\BibTeX{{\rm B\kern-.05em{\sc i\kern-.025em b}\kern-.08em
    T\kern-.1667em\lower.7ex\hbox{E}\kern-.125emX}}

\title{ODES: \underline{O}nline \underline{D}omain Adaptation with \underline{E}xpert Guidance for  Medical Image \underline{S}egmentation}

\author{
\IEEEauthorblockN{
Md Shazid Islam,
Sayak Nag,
Arindam Dutta,
Sk Miraj Ahmed,\\
Fahim Faisal Niloy, Shreyangshu Bera,
Amit K. Roy-Chowdhury, \IEEEmembership{Fellow, IEEE}
}
\IEEEauthorblockA{
Department of Electrical and Computer Engineering, University of California, Riverside \\
\texttt{\{misla048, snag005, adutt020, sahme047, fnilo001, sbera004, amitrc\}@ucr.edu}
}
}

\begin{document}

\maketitle
\thispagestyle{firstpage} % Apply only to this page

\begin{abstract}
Unsupervised domain adaptive segmentation typically relies on self-training using pseudo labels predicted by a pre-trained network on an unlabeled target dataset. However, the noisy nature of such pseudo-labels presents a major bottleneck in adapting a network to the distribution shift between source and target domains. This challenge is exaggerated when the network encounters an incoming data stream in online fashion, where the network is constrained to adapt in exactly one round of forward and backward passes. In this scenario, relying solely on inaccurate pseudo-labels can lead to low-quality segmentation, which is detrimental to medical image analysis where accuracy and precision are of utmost priority. We hypothesize that a small amount of pixel-level annotation obtained from an expert can enhance the performance of domain adaptation of online streaming data, even in the absence of dedicated training data. We call our method ODES: \underline{D}omain Adaptation with \underline{E}xpert Guidance for \underline{O}nline Medical Image \underline{S}egmentation that adapts to each incoming batch of data in an online setup, incorporating feedback from an expert through active learning. However, acquiring pixel-level annotations through active learning for all images in a batch often results in redundant data annotation and increases temporal overhead in online adaptation. To reduce annotation time and make the adaptation process more online-friendly, we further propose a novel image-pruning strategy that selects the most useful subset of images from the current batch for active learning. Our proposed approach outperforms existing online adaptation approaches and produces competitive results compared to offline domain adaptive active learning methods.
\end{abstract}


\section{\textbf{Introduction}}

\begin{figure*}[t]
    \centering
    
    \includegraphics[width=0.75\textwidth]{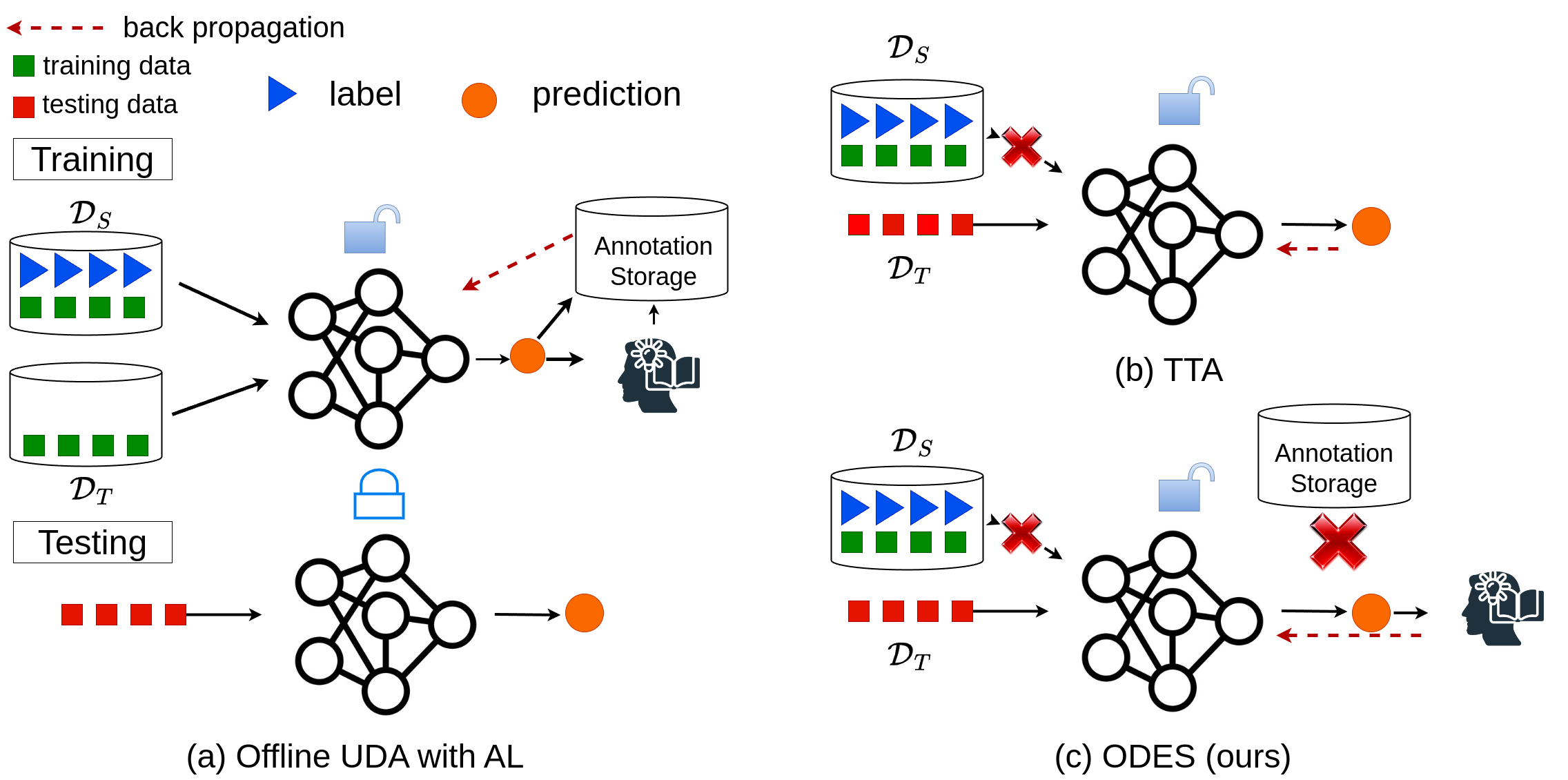}
    \caption{\textbf{Illustration of different domain adaptation setups.} (a) illustrates the offline Active Domain Adaptation (ADA), where labelled source and unlabelled target domain data are used for training and an annotation storage is required to store the annotation from the active learner which is used later part of training. (b) shows test-time adaptation (TTA) setup. (c) illustrates the our proposed setup ODES, where we do not allow any access to source data or any kind of data storage.}
    \label{fig:all_setup}
\end{figure*}

Semantic segmentation is a fundamental task in medical image analysis, enabling the detection of abnormalities in internal organs and the precise study of their shapes, thereby facilitating long-time disease monitoring \cite{liu2022region,campello2021multi} and subsequent treatment \cite{sharma2010automated,lin2013online}.
% Image semantic segmentation is a fundamental task in medical image analysis, enabling as accurate medical image segmentation facilitates treatment \cite{sharma2010automated}, rehabilitation \cite{lin2013online}, and long-time disease monitoring \cite{liu2022region} by analyzing the precise shape of internal organs and detecting abnormalities.
In recent years, deep learning-based models have shown impressive performance in medical image segmentation \cite{asgari2021deep,ronneberger2015u,anisha2015pragmatic}.
% In recent years, deep learning-based medical image segmentation \cite{asgari2021deep,ronneberger2015u} has shown impressive performance in diagnosis and treatment \cite{anisha2015pragmatic}   . 
However, these models require extensive, fully annotated training data to achieve their performance gains, which makes the process time-consuming and expensive \cite{ibtehaz2020multiresunet}.
Furthermore, a model trained on one dataset might exhibit poor performance on another dataset due to domain shift, stemming from variations in imaging equipment and protocols \cite{full2021studying}. In order to address these bottlenecks, Unsupervised Domain Adaptation (UDA) \cite{yang2022source,LE-UDA,li2021generalized} has been proposed as an effective approach in the literature. 
The core idea of UDA is to use labeled data from a source domain along with unlabeled data from the target domain to perform self-training  in order to mitigate domain shifts between two distributions.
% The core idea of UDA is to use labeled data from a source domain along with unlabeled data from the target domain to perform self-training to minimize the distribution discrepancy  . 
However, most existing UDA methods rely on two major assumptions: firstly, access to the \emph{labeled source data} \cite{yang2022source}, which is not always guaranteed due to privacy concerns especially when medical data is used, and secondly, access to large amounts of \emph{unlabeled target domain data} for training the model over multiple epochs in order to adapt it to the new domain. This form of adaptation is called offline adaptation where dedicated training data of target domain is used to align the source and target domains before deploying the model.
% entire training data of the target domain throughout the training process, learning from batches of data across multiple epochs. \cite{lu2022unsupervised}  . 
% These training data are used to adapt the model before deployment, which makes the adaptation procedure offline  .
Nonetheless, in practical healthcare settings, an expert has to deal with a continuous flow of data from various patients with no information about the future patient (online setup), making existing offline UDA approaches ineffective. Furthermore, offline UDA necessitates expensive data storage mechanisms for caching the large-scale training data of a target domain. These issues motivate the development of an adaption method favorable for on-the-fly adaptation to distribution shifts of a streaming data consequently alleviating the need for expensive data storage.

 % In addition, the utilization of data from past patients requires data storage that poses the risk of leakage through hacking, phishing, and other means. 

% TTA enables model adaptation through a single pass of an online test batch, bypassing the need for multiple passes of previous batches, which would require the storing of the batched data, as is the case in standard UDA.

% \textcolor{blue}{Our proposed approach involves the self-training of a source pre-trained model on incoming streams of target domain data.}

Offline UDA methods generally rely upon self-training, and pseudo-label refinement is at the crux of such self-training approaches\cite{yang2022source,LE-UDA,li2021generalized}. However, pseudo labels are very susceptible to class imbalance \cite{wei2021crest}. For example, the largest internal organ, `liver' will certainly incorporate more pixels than small organ `spleen'. Hence, self-training will be biased to segmenting  `liver', resulting in subpar performance in segmenting `spleen'. The noisy nature of these pseudo labels contribute to error propagation leading to poor performance \cite{wang2022continual}. To address this issue, we propose to incorporate Active Learning (AL) in our online adaptation instead of relying on only pseudo-labels. The goal of incorporating AL is to use budgeted manual annotation for samples, the model produces the most noisy pseudo labels depicting the highest levels of model confusion. In the context of semantic segmentation the samples constitute the pixels of an image, and incorporating Al alleviates the adverse effect of error propagation due to inaccurate pseudo labels. AL is particularly feasible for medical image analysis since an expert is usually involved during the medical image-capturing process. AL has been successfully incorporated in offline UDA methods and is formally referred to as Active Domain Adaptation (ADA) \cite{shin2021labor,xie2022towards}.

 % Adaptation on online streaming data requires self-training which relies heavily on pseudo labels. However, pseudo labels are very susceptible to class imbalance \cite{wei2021crest}. For example, the largest internal organ, `liver' will certainly incorporate more pixels than small organ `spleen'. Hence, self-training will be biased to segmenting  `liver', resulting in subpar performance in segmenting `spleen'. Also, inaccurate pseudo labels can contribute to error propagation, adversely impacting the performance of online adaptation \cite{wang2022continual}. In this paper, we address these challenges from a different perspective. Rather than relying completely on pseudo-labels, we incorporate budget manual annotation to a small portion of images where the network is least confident. This approach is known as Active Learning (AL), which is quite feasible in the medical domain, as an expert usually gets involved during the medical image-capturing process. Although incorporating AL in UDA, also referred to as Active Domain Adaptation (ADA) \cite{shin2021labor,xie2022towards}, has shown performance boost  compared to standard UDA, this approach still requires the storage of training data and their acquired annotations. 
% \amit{Explain the difference with TTA better and address the issue of timing like you did in the rebuttal. Have a separate para - this question may come up again. }

Formally we refer to our method as \textbf{ODES}, an AL guided domain adaptation method tailored towards online streaming medical image data. To the best of our knowledge, this is the first attempt at incorporating AL into online domain adaptation, particularly for medical image segmentation. Specifically in the online setup a batch of medical images arrives at an expert who utilizes a source pre-trained model for segmenting each image. The expert annotates a budgeted amount of pixels in each image for which the model's predicted pseudo labels are least confident. The newly acquired annotations along with the pseudo labels of the remaining pixels are used to update the model parameters in order to adapt it to the target domain. This setup resembles real-world scenarios in medical practices, where data storage is constrained and online adaptation \emph{post deployment} is necessitated. This form of adaptation is closely related to Test Time Adaptation (TTA) \cite{he2021autoencoder,hu2021fully,wang2020tent}. A detailed distinction between offline UDA, offline ADA, TTA and our setup is illustrated in Fig \ref{fig:all_setup}.
% To address these concerns, we propose \textbf{ODES}, an adaptation guided by Active Learning setup where we assume medical data arrives to the expert in an online streaming fashion and the expert has the scope of annotating an extremely limited number of pixels \cite{xie2022towards}. To the best of our knowledge, this is the first work incorporating AL in domain adaptation for medical image segmentation in an online setup. In our setup, the model encounters a particular batch of data only once, eliminating the necessity for any data or annotation storage. Our setting is closely related to Test Time Adaptation methods (TTA) \cite{he2021autoencoder,hu2021fully,wang2020tent}. 

Existing works on TTA are primarily motivated for applications like autonomous driving, where real-time output is required. On the other hand, ODES is specifically tailored for medical image analysis, an application where real-time output is generally not a critical requirement. The sequential steps of our overall setup, for each patient, encompass data collection, inference, acquisition, and model update. At the inference stage, the patient is provided with the segmentation result immediately after data collection. The patient experiences no waiting time in this step, as the model's minimal inference time ensures prompt results after data collection. During the acquisition stage, uncertainty-guided AL is incorporated by the expert where waiting time occurs due to the manual annotation process. We note that medical facilities generally include a time interval between imaging sessions for consecutive patients, known as \emph{buffer time}. During this interval, radiologists and their teams utilize the time to perform essential tasks such as preparing the imaging equipment, reviewing and updating patient records, and ensuring that the imaging environment is properly sanitized. In our setting, AL can be considered as a part of buffer time. Thus AL can be seamlessly integrated into a practical medical setup without significantly disrupting existing workflows. After updating the model based on the expert feedback, the model is deployed to analyze the medical data of the next patient, and the same cycle is repeated. 

Reducing waiting time and enhancing convenience are crucial considerations in an online application. As AL involves manual annotation, it is not practically convenient for an expert to annotate a budget portion of pixels in every image of a batch especially while working in an online application. Moreover, not all images within the batch carry equal informative value. Hence, applying AL across all images of test batch results in redundant pixel annotations, thereby escalating both annotation costs and time. This motivates us to develop a novel image-pruning technique which selects the subset of the most informative images from the incoming batch. Thus the image pruning technique makes our approach online application friendly by reducing both annotation time and burden on the expert. The major contributions of this work are summarized as follows:

\begin{itemize}
    \item We are the first to propose a novel AL-guided domain adaptation in online streaming data, \textbf{ODES}, for storage and source-free medical image segmentation. 

    \item We propose a novel image pruning strategy to significantly reduce the annotation burden and the time cost associated with AL.
    
    \item We perform extensive experimentation on publicly available medical image datasets and show that with as little as $1\%$ annotations acquired as active feedback, \textbf{ODES} not only outperforms existing TTA approaches but also reaches near to offline ADA performance.  
\end{itemize}

\section{\textbf{Related Works}}
% \noindent \textbf{Test Time Adaptation:}

\begin{figure*}[t]
  \centering
  \includegraphics[width=\textwidth]{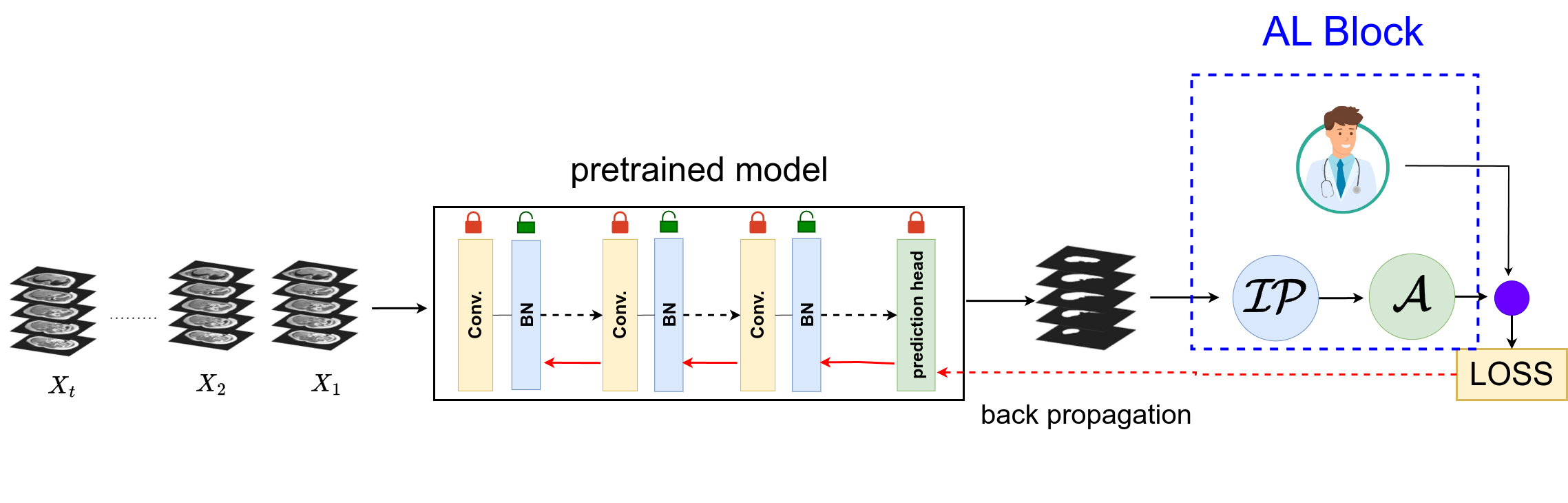} % 
  \caption{The pre-trained model encounters a continuous stream of batched data of target domain. The model first predicts the pseudo-labels of the current batch. Following this, the active learning block (AL block) uses Image Pruning ($\mathcal{IP}$) (Sec \ref{sec:IP}) to prune the test batch and obtains a subset of $K\%$ most informative images in the batch. Next, pseudo labels of the selected images are passed to the Acquisition ($\mathcal{A}$) block (Sec. \ref{sec:PA}) which selects the most uncertain pixels or patches with the budget $b\%$ pixels of in each of these images for annotation acquisition from the active learner (Expert). Next, the Batch Normalization (BN) layers of this model are updated by minimizing a loss function which comprises a supervised loss from annotated pixels/patches from active learning and an unsupervised continuity loss.(Sec. \ref{sec:adapt}).}
  \label{fig:workflow}
\end{figure*}

\subsection{Adaptation on Streaming data}
% \noindent{\textbf{Test Time Adaptation:}} 

  % TTA addresses this issue by adapting the model weights according to the characteristics of the current test batch in the target domain. 

% Being an offline adaptation approach, UDA is not suitable for situations where a continuous flow of data is present and adaptation needs to be done on the fly.
 
 UDA is an offline adaptation which may \cite{LE-UDA} or may not \cite{yang2022source} require the source data. It also requires the entire target domain data to be available during adaptation. If we impose additional constraints, such as streaming data batches of target domain and lack of access to source data, the scenario closely resembles Test Time Adaptation (TTA). While UDA methods are fully based on offline adaptation before deployment for inference, TTA adapts the model post-deployment, during inference or testing. Some TTA methods \cite{wang2020tent,niu2022efficient,BatesonTTA,hu2021fully} solely update the batch normalization layer for adaptation, all employing entropy minimization with distinct information sources like shape moments \cite{BatesonTTA}, contour regularization \cite{hu2021fully}, and reliable, non-redundant sampling \cite{niu2022efficient}.  The method described in \cite{karani2021test} employs denoising autoencoders and updates the parameters of the shallow image normalization CNN in test time adaptation. CoTTA \cite{wang2020tent} employed a teacher-student model where all the weights of the student model are updated using consistency loss and the teacher model is updated using the exponential moving average method. Instead of global domain adaptation to all categories, SATTA \cite{zhang2023satta} updates model parameters for each semantic category individually. Another method SaTTCA \cite{li2023scale} has shown promising result in TTA for segmenting lesions of various sizes. Another method DLTTA \cite{yang2022dltta} introduced a dynamic learning rate adjustment framework for test-time adaptation. However, this method relies on additional data storage as it requires a memory bank to estimate the discrepancy and adaptation demand for each test sample. Adaptation in TTA methods suffers from catastrophic forgetting \cite{hu2019overcoming} which means gradually drifting away from source knowledge. In order to tackle this issue some techniques such as stochastic restoration \cite{wang2022continual} and fisher regularization \cite{kirkpatrick2017overcoming} have been employed.

 % \\
% \noindent{\textbf{Active Learning-Based Domain Adaptation:}}
% There has been lots of work using AL in medical image segmentation \cite{wen2018comparison,yang2017suggestive,9301335}. Most of them rely on sampling functions based on uncertainty \cite{AL_ent2,wang2016cost,shen2018deep}. Recently, ADA methods have gained popularity to mitigate the shortcomings of UDA. RIPU \cite{xie2022towards} designed a sampling strategy based on the uncertainty and regional purity for the segmentation task. Another method, LabOR \cite{shin2021labor} employed a pixel-labeling strategy based on inconsistency mask, while EADA \cite{xie2022active} relied on free energy measurement to perform ADA. However, all of these existing strategies are defined for the offline setup. Our proposed approach addresses this gap in the literature by showing an effective method for integrating AL in a TTA setup.

\subsection{Active Learning-Based Domain Adaptation}
Active learning (AL) is an efficient approach that can reduce the annotation cost while enhancing the performance of deep learning models. In order to maximize the performance of the deep learning model with a budget annotation an effective sampling function is needed.  Most of the sampling function rely on sampling functions based on uncertainty \cite{AL_ent2,wang2016cost,shen2018deep,nath2020diminishing}. The uncertainty can be measured in terms of entropy \cite{joshi2009multi}, Maximum Normalized Log-Probability \cite{shen2018deep}, and Bayesian uncertainty measurement \cite{pmlr-v70-gal17a}. Gaillochet et. al. \cite{gaillochet2023active} proposed a novel batch querying strategy which computes uncertainty at the level of batches instead of individual samples and can be integrated with any uncertainty-based approach.

Annotation for image segmentation task is more expensive and time-consuming than classification task because segmentation involves dense prediction. To address this researchers have incorporated active learning with the standard UDA setup, colloquially referred to as Active Domain Adaptation (ADA). RIPU \cite{xie2022towards} designed a sampling strategy based on the uncertainty and regional purity for the segmentation task. Another method, LabOR \cite{shin2021labor} employed a pixel-labeling strategy based on inconsistency mask, while EADA \cite{xie2022active} relied on free energy measurement to perform ADA. However, all of these existing strategies are defined for the offline setup. Our proposed approach addresses this gap in the literature by showing an effective method for integrating AL in a TTA setup.

\section{\textbf{Methodology}}
We design an active learning-guided TTA framework with application to medical image segmentation. Initially, a segmentation model $f_{\theta}$ is trained on a set of labeled source data  $\mathcal{S}= \{(X_{S}^i ,Y_{S}^i)\}_{i=1}^{M_s} \sim \mathcal{D}_s$ to segment total $C$ number of classes, where $\mathcal{D}_s$ is the source domain data distribution. As shown in Fig. \ref{fig:workflow}, following the TTA setup \cite{wang2020tent} the inference and adaptation process is continuous in nature whereby the model encounters a continuous stream of batches $\mathbf{X}_1 \rightarrow  \mathbf{X}_2 \rightarrow ... \rightarrow \mathbf{X}_t \rightarrow ...$. A batch can be represented by $\mathbf{X}_t = \{X_{\mathcal{T}}^j \}_{j=1}^{\mathcal{B}_t}$ where each $X_{\mathcal{T}}^j \in \mathcal{T}$ with $\mathcal{T}$ being the target domain having a different data distribution $\mathcal{D}_T$.The entire TTA process follows an infer, acquire, and update policy. At first, the model performs inference on the current batch and then, based on our sampling strategy, highlights images and pixels that need to be annotated via active learning. After acquiring the budgeted annotations, the model is updated via supervised empirical risk minimization. This form of active feedback-guided TTA enables the model to improve progressively with each incoming stream of test data.

% In the subsequent sections, we describe our active learning and domain adaptation approach.
\subsection{Active Learning }
\label{sec:Active Learning}
First, the pre-trained model $f_{\theta}$ is used to infer on a batch $\mathbf{X}_t$ and obtain pseudo labels $\mathbf{\hat P}_t = \{ \hat P_j \}_{j=1}^{\mathcal{B}_t}$. The active learner (expert) provides budgeted annotation based on these pseudo-labels. However, all the images do not exhibit the same amount of domain shift in a test batch. We hypothesize that annotations from the active learner can be efficiently utilized if the annotation is spent on the images with larger domain shifts instead of annotating all the images of the test batch. This strategy not only alleviates the annotation burden on the expert, but also reduces the waiting time associated with AL. 
% \amit{Is it possible to draw a small figure as wrapfig to explain IP and PA steps?}
\subsubsection{Image Pruning} 
\label{sec:IP}
% \noindent \textbf{Image Pruning:} 

\begin{figure*}
  \centering
  \includegraphics[width=\textwidth]{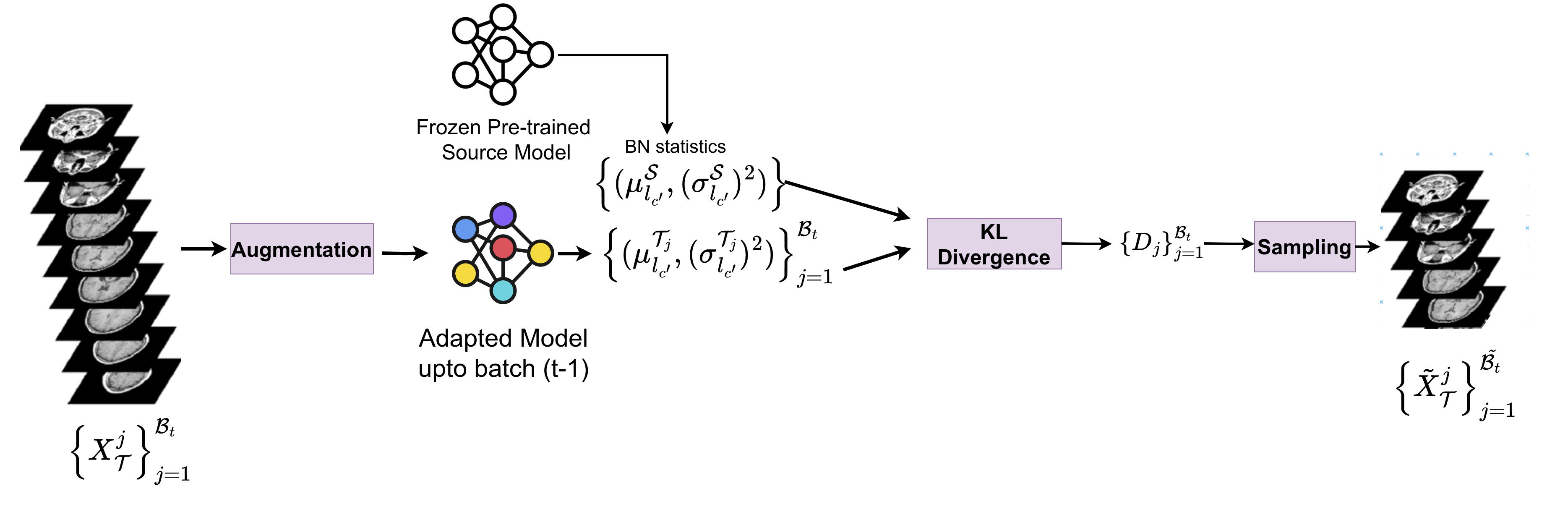} 
  \caption{  This figure illustrates the image pruning strategy. For each image in the batch of the target domain, we perform augmentation and extract feature statistics as the mean and variance from the BN layer. We then compare these statistics against the stored BN statistics of the pre-trained source model using KL divergence. The top $K \%$ of images with the highest KL divergence are selected for use in active learning.}
  \label{image_pruning}
\end{figure*}

To select the images with larger domain shifts, we leverage the batch-normalization layer (BN) statistics of incoming test batches. The statistics of $\mathcal{D}_S$ are stored in the BN layer of the $f_{\theta}$ in terms of running mean and running variance. We compare the feature statistics for each $X_{\mathcal{T}}^j \in \mathbf{X}_t$ with the source statistics. With a domain shift, an abrupt change in feature statistics can be visible in terms of KL divergence \cite{hershey2007approximating}. Therefore, for each $X_{\mathcal{T}}^j$ in the batch we first augment it to obtain $\tilde{X}_{\mathcal{T}}^j$. Assuming the per-channel BN layers to exhibit a Gaussian distribution the divergence between the statistics of $f_{\theta}$ (approximated as ${\cal{N}}(\mu_{l_{c'}}^{\mathcal{S}},(\sigma_{l_{c'}}^{\mathcal{S}})^2)$ and the BN statistics of $\tilde{X}_{\mathcal{T}}^j$ (approximated as ${\cal{N}}(\mu^{\mathcal{T}_{j}}_{l_{c'}},(\sigma^{\mathcal{T}_{j}}_{l_{c'}})^2)$ is defined as
\begin{equation}
\label{eq:slice_dist}
% \footnotesize
D(\mathcal{S},\tilde{X}_{\mathcal{T}}^j) = \sum_{l} \sum_{c'}\text{KL} \left [ \mathcal{N}\left(\mu_{l_{c'}}^{\mathcal{S}},(\sigma_{l_{c'}}^{\mathcal{S}})^2\right) ,  \mathcal{N}\left(\mu^{\mathcal{T}_{j}}_{l_{c'}},(\sigma^{\mathcal{T}_{j}}_{l_{c'}})^2\right)\right] 
\end{equation}
\noindent The higher the value of $D(\mathcal{S},\tilde{X}_{\mathcal{T}}^j)$ the greater the domain shift. Therefore, we select  $K\%$ images from each $\mathbf{X}_t$ with the highest values of $D(\mathcal{S},\tilde{X}_{\mathcal{T}}^j)$ and remove the remaining resulting in a pruned batch, $\tilde{\mathbf{X}}_t$, with batch size $\tilde{\mathcal{B}_t} < \mathcal{B}_t $. Fig. \ref{image_pruning} illustrates the image pruning strategy.
\subsubsection{Acquisition Function} 
\label{sec:PA}

After obtaining the pruned test batch $\tilde{\mathbf{X}}_t$,   $b\%$ budgeted pixels from each of its images are annotated by the active learner. For sampling, we follow the acquisition function proposed in \cite{xie2022towards}. We shall conduct experiment both for pixel-based acquisition and patch-based acquisition. Patch-based acquisition is done for only square shaped patch. For both types of acquisitions, we assess prediction uncertainty and regional impurity. 

\noindent \textbf{Prediction Uncertainty:} 
The entropy ($\mathcal{H}$) of a pixel $(x,y)$ can be computed by
\begin{equation}
\mathcal{H}(x,y)= - \sum_{c}  \mathbf{P}(x,y,c)\log \mathbf{P}(x,y,c)
\label{eq:h}
\end{equation}
$\mathbf{P}$ is the softmax output of the network prediction, $(x,y)$ is the pixel coordinate, and $c$ is the class.

For pixel-based acquisition, the prediction uncertainty ($\mathcal{U}$) is equal to entropy entropy ($\mathcal{H}$) at that pixel.

\begin{equation}
\mathcal{U}(x,y)= \mathcal{H}(x,y)
\label{eq:pu_p}
\end{equation}

For, patch-based acquisition, we consider a square area of $A(x,y)$ with center at $(x,y)$. The prediction uncertainty ($\mathcal{U}$) can be expressed by the average entropy in that area which is expressed in the following equation:

\begin{equation}
\mathcal{U}{(x,y)}=\frac{1}{\left|A(x,y)\right|} \sum_{(u, v) \in A(x,y)} \mathcal{H}{(u,v)}
\label{eq:pu_a}
\end{equation}

\begin{figure}
  \centering
  \includegraphics[width=0.7\columnwidth]{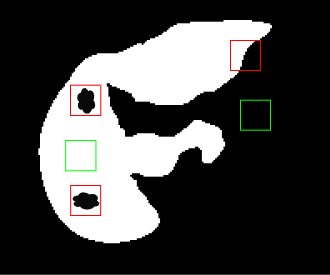} 
  \caption{Regional Impurity shown for different regions. The square regions with green colour do not have any regional impurity as there is no mix of different semantics. According to equation \ref{eq:p} the regional impurity becomes 0 in these areas. The square regions with red colour is impure,because there is a mix of different semantics.}
  \label{region_impurity}
\end{figure}

\noindent \textbf{Regional Impurity:} Regional impurity ($\mathcal{P}$) indicates whether there is a mixture of different semantics in a selected region. If there are mixture of semantics, that region is considered impure. Regional impurity ($\mathcal{P}$) is computed, which indicates whether there is a mixture of different instances in a square area around $(x,y)$ coordinate. Regional impurity can be expressed by

\begin{equation}
\mathcal{P}(x,y) = -\sum_{c=1}^C \frac{\left|A^c(x,y)\right|}{\left|A(x,y)\right|} \log \frac{\left|A^c(x,y)\right|}{\left|A(x,y)\right|}
\label{eq:p}
\end{equation}
\noindent where $\left|A^c(x,y)\right|$ is the number of pixels of class $c$ in the square region around pixel $(x,y)$ and $\left|A(x,y)\right|$ is total area of the square. This equation holds both for pixel-based and patch-based acquisition.

The overall acquisition function for each image in $\tilde{\mathbf{X}}_t$ is given as,
\begin{equation}
% \footnotesize
\mathcal{A}_j(x,y)=\mathcal{U}_j(x,y) \odot \mathcal{P}_j(x,y)
\label{eq:u}
\end{equation}

where $j \in \{1,2, ... , \tilde{\mathcal{B}_t} \}$

For pixel based acquisition, sampled from the image for annotating are given as,
\begin{equation}
\footnotesize
\mathcal{Q}^*=\left\{(x^{*},y^{*})_{j}\right\}_{j=1}^{\tilde{\mathcal{B}_t}} \text{where } 
(x^{*},y^{*})_{j}  =  \left\{ 
\text{Top $b\%$ of }  \underset{(x,y) \in R^{2}} {\arg \max } \text{ } \mathcal{A}_j(x,y)\right\}
% \end{array}
\label{eq:q}
\end{equation}

For patch-based acquisition, sampled from the image for annotating are given as,

% \begin{equation}
% \footnotesize
% \mathcal{Q}^*=\left\{A(x^{*},y^{*})_{j}\right\}_{j=1}^{\tilde{\mathcal{B}_t}} \text{where } 
% (x^{*},y^{*})_{j}  =  \left\{ 
% \text{Top $\frac {b\%}{     \left|A(x,y)\right|  }$ of }  \underset{(x,y) \in R^{2}} {\arg \max } \text{ } \mathcal{A}_j(x,y)\right\}
% % \end{array}
% \label{eq:r}
% \end{equation}

\begin{equation}
\footnotesize
\begin{aligned}
\mathcal{Q}^* &= \left\{A(x^{*},y^{*})_{j}\right\}_{j=1}^{\tilde{\mathcal{B}_t}} \text{ where } \\
(x^{*},y^{*})_{j}  &=  \left\{ 
\text{Top } \frac {b\%}{\left|A(x,y)\right|} \text{ of }  \underset{(x,y) \in R^{2}} {\arg \max } \text{ } \mathcal{A}_j(x,y)\right\}
\end{aligned}
\label{eq:r}
\end{equation}

where $\left|A(x,y)\right|$ indicates area of each square patch. We note that the total amount of pixels annotated by activate learning is the same for both pixel and patch based annotation. There will be a time cost for the expert to provide budget annotations. However, in online applications, a reduced waiting time will increase efficiency by mitigate the gap between input and output data rates. Therefore, by selecting the most informative subset of images from a batch for AL, the waiting time can be significantly reduced.

% There will be a waiting time for the expert to provide budget annotations. However, in online applications, reducing waiting time increases efficiency as there is a continuous flow of data. OASIS allows the expert to focus on a subset of images in each batch, reducing the number of images requiring annotation through AL. Thus, OASIS reduces waiting times and alleviates the burden of annotation.
\subsection{Adapting the Model}
\label{sec:adapt}

After $b\%$ pixels have been selected from each image in $\tilde{\mathbf{X}}_t$, their corresponding labels are acquired from the active learner. To adapt the model $f_{\theta}$ to $\mathcal{T}$, we update the parameters of BN layers by minimizing the supervised cross-entropy loss on the annotated pixels obtained from AL and an unsupervised continuity loss. The supervised loss can be expressed by,
\begin{equation}
% \small
\mathcal{L}_{sup}=-\frac{1}{|\mathcal{Q}^*|} \sum_{(x,y) \in \mathcal{Q}^*} \sum_{c=1}^C \mathbf{Y}(x,y,c) \log \mathbf{P}{(x,y, c)}
\label{loss_sup}
\end{equation}
where $\mathbf{Y}(x,y,c)$ is the label provided by active learner at the $(x,y)$ pixel for class $c$. 

We note that a batch of data comprises a series of two-dimensional images, which are then combined to form a three-dimensional volumetric stack representation. The images within the three-dimensional stack are arranged in a continuous manner, ensuring a smooth transition between adjacent slices. From this observation, we formulate an unsupervised continuity loss ($\mathcal{L}_{cont}$), which minimizes any abrupt changes between successive two-dimensional images. $\mathcal{L}_{cont}$ can be expressed by

\begin{equation}
\mathcal{L}_{cont}= \sum_{j=1}^{\mathcal{B}-1} CE(\mathbf{\hat P}^j,\mathbf{\hat P}^{j+1} )
\label{loss_unsup}
\end{equation}

where $\mathcal{B}$ is the batch size, $\mathbf{\hat P}^j$ is the pseudo-label of the $j-th$ image of the batch and CE is the cross-entropy loss. The total loss function becomes

\begin{equation}
\mathcal{L}_{total}= \mathcal{L}_{sup} + \lambda \mathcal{L}_{cont} 
\label{loss_total}
\end{equation}

% negative learning loss \cite{kim2019nlnl}.

% \begin{equation}
% \pi\left(\mathbf{P}^{(x,y, c)}\right)= \begin{cases}1 & \text { if } \mathbf{P}^{(x,y, c)}<\tau \\ 0 & \text { otherwise }\end{cases}
% \end{equation}

\begin{algorithm}[t]
\footnotesize
\caption{\textbf{ODES} \\
Image Selection Rate $K \%$ per batch, AL budget $b \%$ pixel per image}
\label{alg:Drfrodo}
\begin{algorithmic}[1]
\footnotesize
\REQUIRE Source pre-trained model $f_{\theta}$,
\STATE calculate $\left\{(\mu_{l_{c'}}^{\mathcal{S}},(\sigma_{l_{c'}}^{\mathcal{S}})^2)\right\}$ which is the BN statistics of $f_{\theta}$.
 % \in \left\{1,2,3,... , N \right\}
    \FOR{each batch $t = 1,2, \cdot \cdot \cdot$}

        \STATE Initialize divergence array $\text{DIV} = \left\{ \right\}$ 
        \FOR{each image $j \in \left\{X_{\mathcal{T}}^j  \right\}_{j=1}^{\mathcal{B}_t}$ }
            \STATE  $X_{\mathcal{T}}^j$ $\leftarrow$ Augmentation
            \STATE  Calculate BN statistics $\left\{(\mu^{\mathcal{T}_{j}}_{l_{c'}},(\sigma^{\mathcal{T}_{j}}_{l_{c'}})^2)\right\} $ of augmented $X_{\mathcal{T}}^j$
            \STATE Calculate $D$ using equation \ref{eq:slice_dist}
            \STATE DIV $\leftarrow$ Store $D$
            
        \ENDFOR
    \STATE $\tilde{\mathbf{X}}_t$ $\leftarrow$ Select images with top-$K \%$ value in DIV
    \STATE   perform pixel sampling on $\tilde{\mathbf{X}}_t$ with budget $b \%$ using equation \ref{eq:pu_p},\ref{eq:p},\ref{eq:u},\ref{eq:q} Or, perform patch sampling on $\tilde{\mathbf{X}}_t$ with budget $b \%$ using equation \ref{eq:pu_a},\ref{eq:p},\ref{eq:u},\ref{eq:r} 
    \STATE update model using equation \ref{loss_sup}, \ref{loss_unsup}, \ref{loss_total}.
    \ENDFOR
\end{algorithmic}
\end{algorithm}

\section{\textbf{Experiments and Results}}

\begin{table*}[t]

\centering
% \footnotesize

\caption{Comparison among different TTA Methods reported in terms of DSC. Pixel AL and Patch AL indicate pixel and patch-based active learning, respectively. The best results are highlighted in \textcolor{red}{red} and the second best results in \textcolor{blue}{blue}. Here we consider $b=1$. All experiments are done using our own pre-trained model considering only one forward pass of each batch of data.}
\label{tab:TTA}

\begin{tabular}{c| 
>{\centering\arraybackslash}m{12mm} 
>{\centering\arraybackslash}m{15mm}
>{\centering\arraybackslash}m{15mm}
>{\centering\arraybackslash}m{15mm} >{\centering\arraybackslash}m{15mm}|
>{\centering\arraybackslash}m{15mm}|
>{\centering\arraybackslash}m{15mm}}
\hline
 & \multicolumn{5}{c|}{CHAOS T1 (IP $\rightarrow$ OOP)} & CHAOS $\rightarrow$ DUKE & BMC $\rightarrow$RUNMC \\
\hline
 Methods & Liver & L.Kidney & R.Kidney & Spleen & Mean & Liver & Prostate \\
\hline
Source only & 87.77 & 37.97 & 18.92 & 67.31 & 52.99 & 26.28 & 65.32 \\

TENT \cite{wang2020tent} & 86.03$\pm$0.16  & 58.1$\pm$0.15 & 54.72$\pm$0.10 & 70.34$\pm$0.12 & 67.30$\pm$0.13 & 46.64 $\pm$1.47 & 71.02$\pm$0.21\\

CoTTA \cite{wang2022continual} & 86.38$\pm$0.08 & 52.8$\pm$0.10 & 58.27$\pm$0.08 & 71.06$\pm$0.15 & 67.13$\pm$0.10 & 52.58 $\pm$0.80 & 73.87$\pm$0.15\\

F-TTA \cite{hu2021fully} & 86.91$\pm$0.17 & 61.99$\pm$0.11 & 62.11$\pm$0.14 & 69.51$\pm$0.14 & 70.13$\pm$0.14 & 48.29 $\pm$0.68& 73.18$\pm$0.18\\
SaTTCA \cite{li2023scale}& 87.90$\pm$0.08 & 64.85$\pm$0.11 & 65.47$\pm$0.09 & 74.51$\pm$0.11 & 73.19$\pm$0.09 & 58.39 $\pm$0.72 & 74.91$\pm$0.30 \\

% DLTTA \cite{yang2022dltta}&  &  &  &  &  &  & \\

TTAS \cite{BatesonTTA}& 86.74$\pm$0.21 & 61.68$\pm$0.16 & 58.13$\pm$0.17 & 71.07$\pm$0.15 & 69.41$\pm$0.17 & 49.39 $\pm$1.19 & 72.81$\pm$0.22 \\

\hline
ODES (Pixel AL) &  &  &  &  &  &  &  \\

K = 100 & \textcolor{red}{89.15$\pm$0.06}  & \textcolor{red}{73.49$\pm$0.05} & \textcolor{red}{74.02$\pm$0.05} & \textcolor{red}{78.67$\pm$0.03} & \textcolor{red}{78.83$\pm$0.04}  & \textcolor{red}{72.21$\pm$0.43} & \textcolor{red}{79.96$\pm$0.11}\\

K = 50 & \textcolor{blue}{89.09$\pm$0.03}  & \textcolor{blue}{72.46$\pm$0.06} & \textcolor{blue}{73.93$\pm$0.05} & \textcolor{blue}{78.62$\pm$0.06}  & \textcolor{blue}{78.53$\pm$0.05} & \textcolor{blue}{71.32$\pm$0.51}  & \textcolor{blue}{79.10$\pm$0.07} \\

K = 10 & 87.96$\pm$0.06 & 72.27$\pm$0.04 & 69.58$\pm$0.05 & 77.43$\pm$0.06 & 76.81$\pm$0.05 & 66.45$\pm$0.56 & 77.73$\pm$0.08 \\
\hline

ODES (Patch AL) &  &  &  &  &  &  &  \\

K = 100 & \textcolor{red}{88.75$\pm$0.04} & \textcolor{red}{72.38$\pm$0.07} & \textcolor{red}{74.64$\pm$0.05} & \textcolor{red}{78.61$\pm$0.05} & \textcolor{red}{78.62$\pm$0.05} & \textcolor{red}{71.97$\pm$0.49} & \textcolor{red}{79.34$\pm$0.07} \\

K = 50 & \textcolor{blue}{88.42$\pm$0.03} & \textcolor{blue}{72.20$\pm$0.06} & \textcolor{blue}{74.43$\pm$0.04} & \textcolor{blue}{78.38$\pm$0.06} & \textcolor{blue}{78.43$\pm$0.04} & \textcolor{blue}{70.99$\pm$0.41} & \textcolor{blue}{78.36$\pm$0.11} \\

K = 10 & 87.47$\pm$0.05 &  70.77$\pm$0.04 &  69.45$\pm$0.03&  76.61$\pm$0.03& 76.08$\pm$0.04  & 65.90$\pm$0.48  &  77.47$\pm$0.10 \\ 

\hline

\end{tabular}

\end{table*}

\subsection{Dataset}
$\bullet$ \textbf{CHAOS MRI:}
CHAOS (Combined Healthy Abdominal Organ Segmentation) \cite{kavur2021chaos} includes Magnetic Resonance Imaging (MRI) of 20 subjects. MRI data sets are collected for two different sequences which are T1-DUAL and T2-SPIR. T1-DUAL also comprises two different types of signals: in-phase (IP) and out-of-phase (OOP). All the MRI datasets contain the annotations of four human internal organs: liver, left Kidney, right kidney and spleen.

$\bullet$ \textbf{DUKE MRI:}The Duke liver dataset \cite{macdonald_2022_6328447} contains the data of 105 patients. This dataset exhibits four distinct forms of contrasts, namely in phase, opposed phase, T1-weighted, and enhanced T1-weighted. In this dataset, only liver is annotated manually.

$\bullet$ \textbf{Prostate MRI:}A publicly available MRI dataset \cite{liu2020ms}  has been used for prostate segmentation. It comprises manually annotated T2-weighted MRI from two diﬀerent sites: Boston Medical Center (BMC) and Radboud University Medical Center (RUNMC). Each of the datasets comprises 30 MRI stacks.

\subsection{Adaptations}

$\bullet$ \textbf{CHAOS T1-DUAL IP to OOP} 
Although gradient echo sequences of IP and OOP of CHAOS T1-DUAL are collected with the same repetition time, the echo time values \cite{ramalho2012phase} are different causing the domain shift. IP and OOP are considered as source and target domains respectively.

$\bullet$ \textbf{CHAOS T2-SPIR to DUKE} 
 T2-SPIR of CHAOS dataset is T2-weighted. DUKE dataset contains four different contrasts of MRI images. Hence, there is a domain shift between these two datasets. T2-SPIR of CHAOS data is considered the source domain, and DUKE dataset is considered as the target domain for liver segmentation.

% respectively for liver, left kidney, right kidney and spleen segmentation. 
$\bullet$ \textbf{BMC to RUNMC}
BMC is used as the source, and RUNMC dataset as the target domain. 

\subsection{Implementation Details}
We use ResNet-18 backbone in Deeplabv3 \cite{chen2017deeplab}. We have used $1\% (b = 1 ) $ pixel annotation from the active learner in each test batch. For updating the batch normalization layer we have used ADAM  \cite{kingma2014adam} optimizer with a learning rate $0.01$. The input batch of images are normalized using zero mean and unit variance normalization. For patch-based active learning, we consider $ 5 \text{pixel} \times 5 \text{pixel} $ square as a unit region for annotation.

% \begin{table}
% \centering
% \caption{Comparison with Offline Method}
% \label{tab:offline}
% % \begin{tabular}{c|c|c|c|c|c|c|c}

% \hline
%  & \multicolumn{5}{c|}{CHAOS T1} & CHAOS & BMC \\

%  & \multicolumn{5}{c|}{IP $\rightarrow$ OOP} & T2 $\rightarrow$ DUKE & $\rightarrow$RUNMC \\
% \hline
%  Methods & Liver & L. Kid & R. Kid. & Spleen & Mean & Liver & Prostate \\
% \hline
% RIPU & \textbf{93.52} & \textbf{87.42} & \textbf{86.71} & \textbf{86.95} & \textbf{88.65} & \textbf{80.12} & \textbf{89.01} \\

% RIPU-SF & 93.54 & 85.69 & 84.29 & 86.69 & 87.55 & 78.48 & 88.16 \\
% \hline
% DrONDA &  &  &  &  &  &  &  \\

% K = 100 & 90.84 & 83.59 & 85.24 & 81.29 & 85.24 & 77.72 & 85.12 \\

% K = 50 & 91.04 & 85.63 & 82.42 & 80.61 & 84.93 & 76.94 & 83.49 \\

% K = 10 & 90.63 & 78.98 & 82.16 & 79.83 & 82.9 & 75.46 & 80.31 \\
% \hline

% \end{tabular}

% \end{table}
\begin{table*}
\centering
\caption{Comparison with Offline ADA Method under same annotation budget ($b=1$). The offline ADA method consider multiple forward passes during offline training. ODES considers only one forward pass during adaptation.  }
\label{tab:offline}
% \begin{tabular}{c|c|c|c|c|c|c|c}
\begin{tabular}{c| >{\centering\arraybackslash} >{\centering\arraybackslash}m{13mm} >{\centering\arraybackslash}m{13mm} >{\centering\arraybackslash}m{13mm} >{\centering\arraybackslash}m{13mm} >{\centering\arraybackslash}m{13mm}|
>{\centering\arraybackslash}m{15mm}|
>{\centering\arraybackslash}m{15mm}}

\hline
 & \multicolumn{5}{c|}{CHAOS T1 (IP $\rightarrow$ OOP)} & CHAOS $\rightarrow$ DUKE & BMC $\rightarrow$RUNMC \\

\hline
 Methods & Liver & L.Kidney & R.Kidney & Spleen & Mean & Liver & Prostate \\
\hline
RIPU \cite{xie2022towards} & 93.52 & \textbf{87.42} & \textbf{86.71} & \textbf{86.95} & \textbf{88.65} & \textbf{80.12} & \textbf{89.01} \\

RIPU-SF \cite{xie2022towards} & \textbf{93.54} & 85.69 & 84.29 & 86.69 & 87.55 & 78.48 & 88.16 \\
\hline
ODES (Pixel AL) &  &  &  &  &  &  &  \\

K = 100 & 91.41 & 83.73 & 85.46 & 79.81 & 85.04 & 78.11& 84.65 \\

K = 50 & 91.39 & 81.23 & 85.13 & 79.78 & 84.38 & 77.42& 82.86 \\

K = 10 & 90.78 & 80.57 & 82.34 & 77.07 & 82.69 & 76.57& 80.43 \\
\hline

ODES (Patch AL) &  &  &  &  &  &  &  \\

K = 100 & 90.21 & 82.41 & 84.86 & 80.49 & 84.49 & 77.53 & 84.41 \\

K = 50 & 90.47 & 78.93 & 82.95 & 79.76 & 83.03 & 76.76 & 82.84 \\

K = 10 & 89.67 & 74.66 & 82.34 & 79.29 & 81.49 & 75.44 & 79.45 \\
\hline

\end{tabular}

\end{table*}

\subsection{Comparison with other methods}

\subsubsection{Comparison with online adaptation methods}
As our problem setting is closely related to TTA, our baselines are some widely used state-of-the-art (SOTA) TTA methods such as TENT \cite{wang2020tent}, CoTTA \cite{wang2022continual} and SaTTCA \cite{li2023scale}. Dice Score (DSC) is used as the performance metric. In TABLE \ref{tab:TTA}, we have shown the performance comparison, for both pixel-based and patch-based active learning. From TABLE \ref{tab:TTA} we observe that our method has outperformed all other TTA methods in all three adaptations. $K=100$ indicates AL is used in images of test batch, implying no image pruning is involved. When image pruning is involved with $K = 50$ which means only $50\%$ images of the test batch are annotated using AL, we observe only $0.3\%, 0.89\% \text{ and } 0.86\%$ performance drop in all adaptations in pixel-based active learning. Even though annotation cost was reduced by $50 \%$, there was a very insignificant reduction in performance, which indicates the efficacy of our image pruning strategy. Also, for $K = 10$ which means annotation was reduced by $90 \%$, compared to $K = 100$, we observed drop of performance by $2.02 \% , 5.76 \% , 2.23 \%$ in all three adaptations, which is very small amount compared to reduction of annotation  cost. In patch-based active learning we also observe the same pattern. In addition, we note that adding very minimal annotation to TTA can significantly boost performance, specially in CHAOS $\rightarrow$  DUKE where wider range of variations (four distinct forms of contrasts) is seen in target domain. From Table \ref{tab:TTA}, in CHAOS $\rightarrow$ DUKE adaptation, ODES outperforms the best-performing TTA method SaTTCA \cite{li2023scale} by $11.46 \%$ and $15.93 \%$ for K = 10 and 50, respectively.

% CHAOS->DUKE adaptation encounters a wider range of variations (four distinct forms of contrasts) in the target domain (supplementary: section 1.1)

% \subsection{Comparison with offline method}
\subsubsection{Comparison with offline adaptation methods}

\begin{figure*}
    \centering
    \includegraphics[width=\textwidth]{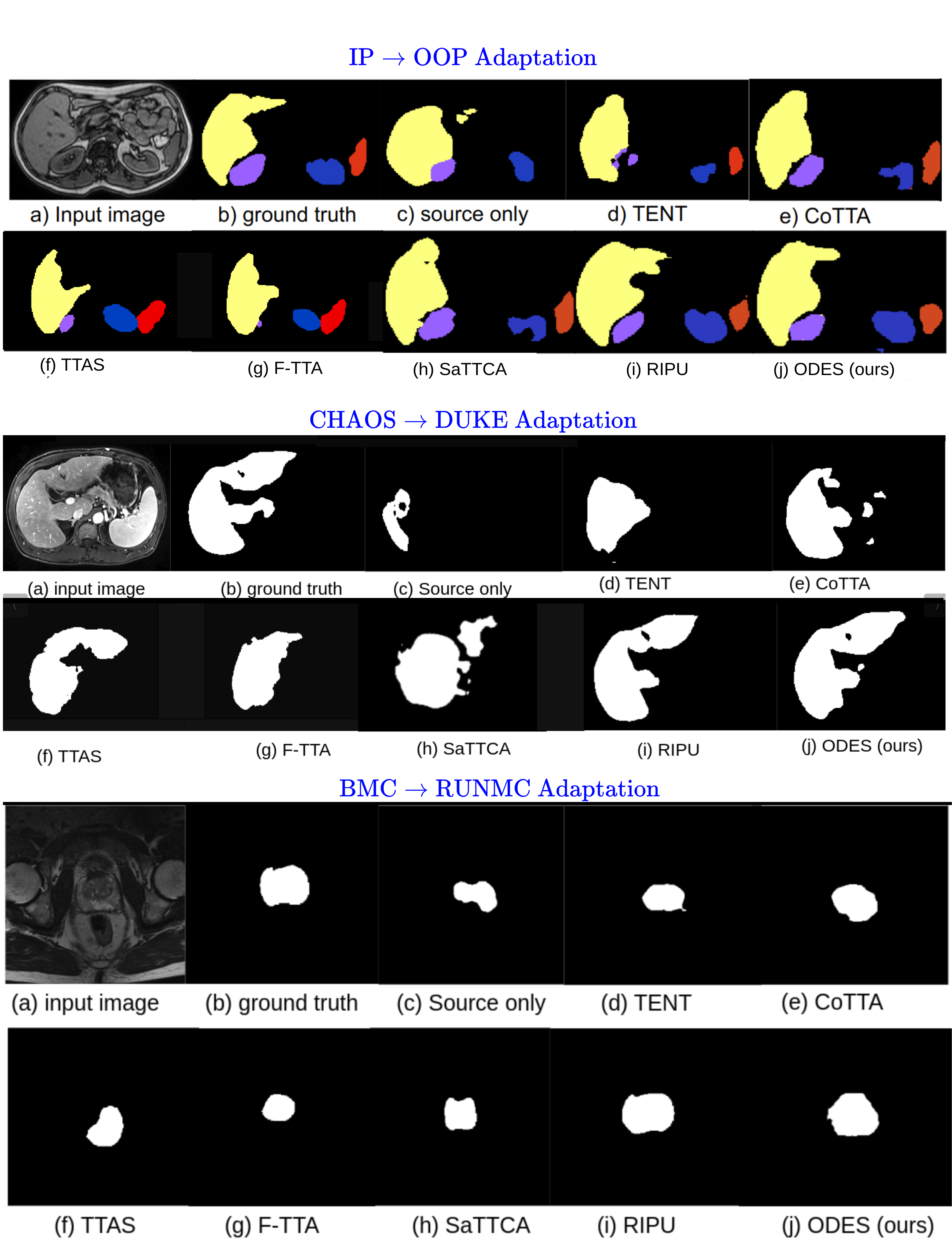}
    \caption{Qualitative Results for IP $\rightarrow$ OOP , CHAOS $\rightarrow$ DUKE and BMC $\rightarrow$ RUNMC adaptation.}
    \label{fig:output_T1}
\end{figure*}

We compare ODES with ADA method RIPU \cite{xie2022towards}. Two variations of RIPU are considered: with source (standard RIPU) and without access to source data (RIPU-SF). Being an offline method, RIPU requires a training set. So we split the entire target data into training ($80 \%$) and testing ($20 \%$) split. As RIPU is an offline method, all the training data is considered to be available together during training and the model can learn from same batch of data through multiple epochs. In ODES, we do not consider that all the training data are available together, rather each batch of data encounters the model in a streaming manner and model is allowed to learn from a batch of data only using a single pass. For this particular experiment, ODES will incorporate active learning only for the batches which is in the training split used in RIPU. In the test split, no active learning is involved. For fair comparison, all results are reported on test split in TABLE \ref{tab:offline}. ODES is a source-free approach and it has a performance gap of $2.51 \% , 0.37 \% \text{ and }, 3.51 \%$ for pixel-based AL and $3.06 \% , 0.95 \% \text{ and }, 3.75 \%$ for patch-based AL  compared to RIPU-SF for $K =100$ in all adaptations. The performance gap occurred as ODES adapts only by a single pass of a particular batch of images, unlike offline RIPU, RIPU-SF where multiple passes are allowed. Despite being an online method with a single pass constraint, ODES performs closely to the offline ADA method. Fig. \ref{fig:output_T1} illustrates the visual comparison among different methods. 

% \subsubsection{Analysis on Different Patch Size }
% In TABLE \ref{tab:patch_size}, we compare the performance of patch based AL for patches of different sizes while keeping the total budget same. We observe that the performance gets better when the size of the patch gets smaller. In patch based annotation, the active learner is required to annotate a region where not all pixels are at least confidence levels for the model. So the active learner does not always annotate the most informative pixels which results in sub-optimal performance. Hence, we observe a drop of performance $1.93 \%$ in mean DSC when patch size was increased from $3 \times 3$ to $11 \times 11$ .

\subsection{Ablation Studies}

\subsubsection{Comparison on Pixel-based and Patch-based Active Learning}
% \amit{I do not like Comments on. I would prefer a subsection called Discussion of Experimental Results and then put E and G under there as subsubsections. Bring Ablation Studies above this.}
\begin{table}
\centering
\caption{Comparison of performance for different patch sizes  in IP $\rightarrow$ OOP for $K = 10$.}
\begin{tabular}{c|
>{\centering\arraybackslash}m{5mm}
>{\centering\arraybackslash}m{5mm}
>{\centering\arraybackslash}m{5mm}
>{\centering\arraybackslash}m{5mm}
>{\centering\arraybackslash}m{5mm}}
\hline
Patch size & Liver & L.Kid. & R.Kid. & Spleen & Mean \\
\hline
$3 \times 3$  & 87.64& 70.86& 69.68& 76.52& \textbf{76.18}\\
$5 \times 5$  & 87.47& 70.77& 69.45& 76.61& 76.08\\
$7 \times 7$  & 87.11& 68.95& 69.46& 76.31& 75.46\\
$9 \times 9$  & 87.16& 66.85& 66.70& 76.28& 74.25\\
$11 \times 11$  & 86.96& 67.55& 66.28& 75.81& 74.15\\
\hline
\end{tabular}
\label{tab:patch_size}
\end{table}

 In TABLE \ref{tab:TTA} and \ref{tab:offline}, we compare the performance of AL between pixel and patch-based AL under the same amount of budget per image.  We observe that pixel-based AL performs better than patch-based annotation. Under the same budget, the pixel based AL only annotates the pixels where the model is least confident. However, in patch based AL, the active learner is required to annotate a region where not all pixels are at least confidence levels for the model, resulting in sub-optimal use of budget pixels. Hence, we observe a slight drop of performance in mean DSC for patch-based annotation. In addition, TABLE \ref{tab:patch_size} shows the comparison of the performance of patch-based AL for patches of different sizes while keeping the total budget same. We observe that the performance gets worse when the size of the patch gets larger. This is because when the size of patch becomes larger, the chance of including less informative pixels in the patches gets higher. So the annotation budget for AL is used in a less efficient manner when the size of the patches gets larger. Hence, we observe a drop of performance $2.03 \%$ in mean DSC when patch size was increased from $3 \times 3$ to $11 \times 11$ . Although patch based AL is more user-friendly compared to pixel-based AL as patches are easier to annotate, there is a trade-off between convenience and accuracy.

\subsubsection{Impact of Random Image Pruning}

\begin{table}
\footnotesize
\centering
\caption{Comparison between proposed batch pruning strategy and random batch pruning with $K=10$ in IP $\rightarrow$ OOP adaptation .}
\begin{tabular}{c|
>{\centering\arraybackslash}m{5mm}
>{\centering\arraybackslash}m{5mm}
>{\centering\arraybackslash}m{5mm}
>{\centering\arraybackslash}m{5mm}
>{\centering\arraybackslash}m{5mm}}
\hline
Image Pruning Strategy & Liver & L.Kid. & R.Kid. & Spleen & Mean \\
 \hline
Proposed  & 87.96 & 72.27 & 69.58 & 77.43 & \textbf{76.81}\\
Random    & 87.77 & 68.76 & 69.88 & 76.30 & 75.16\\
\hline
\end{tabular}
\label{tab:random}
\end{table}

ODES has an image pruning strategy, which selects only the images with the largest domain shifts in a batch for AL. TABLE \ref{tab:random} shows if same number of images from the test batch are selected randomly, there is be a performance drop of $1.37 \%$, which indicates the effectiveness of our batch-pruning strategy.

\subsubsection{Reducing $K$ gradually}
% \begin{table}

% \centering
% \caption{Comparison of DSC performance between $K=100$ and $K$ with exponential decay.}
% \label{tab:exponential_decay}

% \begin{tabular}{c| 
% >{\centering\arraybackslash}m{13mm} 
% >{\centering\arraybackslash}m{13mm}
% >{\centering\arraybackslash}m{13mm}
% >{\centering\arraybackslash}m{13mm} >{\centering\arraybackslash}m{13mm}|
% >{\centering\arraybackslash}m{15mm}|
% >{\centering\arraybackslash}m{15mm}}
% \hline
%  & \multicolumn{5}{c|}{CHAOS T1 (IP $\rightarrow$ OOP)} & CHAOS $\rightarrow$ DUKE & BMC $\rightarrow$RUNMC \\
% \hline
% K & Liver & L.Kidney & R.Kidney & Spleen & Mean & Liver & Prostate \\
% \hline
% 100 & 89.21 & 72.43 & 73.24 & 78.71 & 78.40 & 72.33 & 80.77 \\
% \hline
% Exponential decay & 88.12	& 73.20	& 74.01	&78.48	&78.24 & 71.01 & 79.81\\
% \hline

% \end{tabular}

% \end{table}

\begin{table}
\centering
\caption{Comparison between $K=100$ and $K$ with exponential decay in IP $\rightarrow$ OOP.}
\begin{tabular}{c|
>{\centering\arraybackslash}m{5mm}
>{\centering\arraybackslash}m{5mm}
>{\centering\arraybackslash}m{5mm}
>{\centering\arraybackslash}m{5mm}
>{\centering\arraybackslash}m{5mm}}
\hline
K & Liver & L.Kid. & R.Kid. & Spleen & Mean \\
\hline
100 & 89.15  & 73.49 & 74.02 & 78.67 & 78.83  \\
% \hline
Exponential decay & 88.12	& 73.20	& 74.01	&78.48	&78.24\\
\hline
\end{tabular}
\label{tab:exp_decay}
\end{table}

\begin{table*}[h]
% \footnotesize

\caption{Catastropic Forgetting analysis. Mean DSC is reported for IP $\rightarrow$ OOP adaptation for each batch individually for multiple cycles.} 
\begin{center}
% \begin{tabular}{ c||c|c|c|c|c|c } 
\begin{tabular}{ c||
>{\centering\arraybackslash}m{15 mm}
>{\centering\arraybackslash}m{15mm}
>{\centering\arraybackslash}m{15mm}
>{\centering\arraybackslash}m{15mm}
>{\centering\arraybackslash}m{15mm}
>{\centering\arraybackslash}m{15mm}}

\hline
& \multicolumn{6}{c}{Incoming batch id $\rightarrow$}\\
\hline
 cycle &1 &2 &3 & ... & ... & 20 \\ 
\hline
 1  & 67.71 & 82.37 & 79.91 & ... & ... & 82.11 \\
 2  & 77.51 & 87.37 & 87.24 & ... & ... & 88.26 \\
 3  & 80.57 & 88.29 & 89.07 & ... & ... & 89.63 \\
 
\hline

% fixed &77.76 &78.44 &79.81 &83.56 &83.38 &85.53\\
% decreasing &77.76 &78.36 &79.77 &83.38 &83.36 &84.74\\
\end{tabular}
\end{center}
\label{table:froget}

\end{table*}

This experiment begins with a high initial value of $K$ ($K = 100$) and gradually decreases $K$ in exponential decay manner with each batch passing until it reaches approximately 0. It means no image is annotated after the model gets adapted using a certain number of batches. Here we stopped annotation after $60 \%$ of total batches gets adapted. We observe in TABLE \ref{tab:exp_decay} that even though we reduce $K$ exponentially, the performance drop is not much. It indicates that after adapting to a number of batches, the model gets well adapted to the target domain and performs well in the later batches even if no AL is involved. This experiment shows the scope of early stopping AL which can mitigate the concern of waiting time associated with AL.

\subsubsection{Analysis on Catastrophic Forgetting}
We employ a cyclic evaluation to analyze whether catastrophic forgetting occurs in our approach. After one cycle of adaptation of all batches, we repeat the process with the adapted model. In TABLE \ref{table:froget} We observe increased DSC on the same batch of images. We observe that performance has enhanced in the second and third cycles of evaluation. If there was a catastrophic forgetting, the performance would have degraded. The enhanced DSC indicates that catastrophic forgetting did not occur.

% mitigating the the waiting time associated with AL.

% \subsubsection{Impact of Image Selection Factor \textbf{K}}

% The novelty of ODES has two key aspects. Firstly, incorporate active learning (AL) in a  TTA setup. Secondly, a novel image pruning technique to reduce annotation burden and waiting time during AL.
\subsubsection{Analysis on Different Acquisition Functions }
\begin{table}
\centering
\caption{Comparison between different Acquisition function in IP $\rightarrow$ OOP with $K=10$.}
\begin{tabular}{c|
>{\centering\arraybackslash}m{5mm}
>{\centering\arraybackslash}m{5mm}
>{\centering\arraybackslash}m{5mm}
>{\centering\arraybackslash}m{5mm}
>{\centering\arraybackslash}m{5mm}}
\hline
Ac. Func. & Liver & L.Kid. & R.Kid. & Spleen & Mean \\
\hline
Random & 87.55 & 69.41 & 68.31 & 74.71 & 74.92 \\

ENT \cite{shen2017deep} & 87.97	& 70.88 & 69.35	& 75.15	& 75.83\\

SCONF \cite{culotta2005reducing} & 87.91	& 71.22	& 69.52	&75.74	&76.09\\

R.I.P.U. \cite{xie2022towards} & 87.96 & 72.27 & 69.58 & 77.43 & \textbf{76.81}\\

\hline
\end{tabular}
\label{tab:ac. func.}
\end{table}
 
In ODES, we have developed a novel image pruning technique to reduce annotation burden and waiting time during AL. This image pruning technique can be integrated with any acquisition function. In our experiments, we have used the acquisition function described in \cite{xie2022towards} due to its superiority in acquiring the most informative pixels to be annotated by any expert, irrespective of their medical specialties. In TABLE \ref{tab:ac. func.}, we have shown the comparison among different acquisition functions.

\subsubsection{Distance Metric for Image Pruning Strategy}

\begin{table}
\centering
\caption{Comparison between distance metrics for image pruning in CHAOS $\rightarrow$ DUKE (liver segmentation) and BMC $\rightarrow$ RUNMC (prostate segmentation)   for $K = 10$.}
\begin{tabular}{c|
>{\centering\arraybackslash}m{25mm}|
% >{\centering\arraybackslash}m{20mm}
% >{\centering\arraybackslash}m{5mm}
% >{\centering\arraybackslash}m{5mm}
>{\centering\arraybackslash}m{25mm}}
\hline
% Metric & Liver & L.Kid. & R.Kid. & Spleen & Mean \\
% \hline
% $l_1$ & 89.21 & 72.43 & 73.24 & 78.71 & 78.40 \\

% $l_2$  & 88.12	& 73.20	& 74.01	&78.48	&78.24\\

% KL-div & 88.12	& 73.20	& 74.01	&78.48	&78.24\\

Metric & CHAOS $\rightarrow$ DUKE & BMC $\rightarrow$ RUNMC  \\
\hline
$l_1$ & 56.21 & 76.33\\

$l_2$  & 57.12 & 76.51\\

KL-div & \textbf{65.46}	 & \textbf{77.47}\\

\hline
\end{tabular}
\label{tab:dist_pruning}
\end{table}

ODES uses KL divergence to identify the images of the input batch with the highest domain shift. In TABLE \ref{tab:dist_pruning}, we compare KL divergence with other distance metrics for image pruning strategy. As KL divergence has the superior ability to identify the domain shift, it performs better than the other metrics.

\subsubsection{Impact of Annotation Budget ($b$)}
\begin{figure}[t]
  \centering
  \includegraphics[width=0.9\columnwidth]{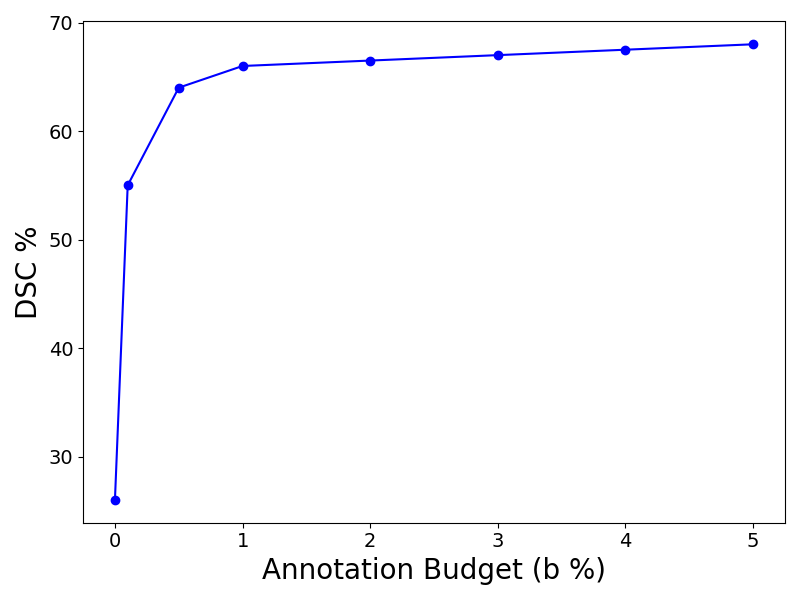} % 
  \caption{Impact of Annotation Budget ($b$) in CHAOS $\rightarrow$ DUKE adaptation for $K=10$. At first, we observe a drastic improvement of DSC with small change of $b$. Later DSC values follow a saturation trend with $b$ increases}
  \label{ablation_b}
\end{figure}

In Fig. \ref{ablation_b}, we show the impact of changing the annotation budget per image $b$. We observe that, initially, there is a drastic improvement of DSC performance even with small change of $b$. Later the DSC reaches saturation. 
It occurs due to the fact that the model is adapted using a single pass of a particular batch. After a certain value of $b$, even though more annotations are involved the model can not perform better due to single pass constraint, which results in the saturation trend. In more forward and backward pass is allowed, the performance can enhance as $b$ increases.

% \subsubsection{Other TTA methods with Active Learning}

% \subsection{Discussion of Experimental Results}

% \subsubsection{Analysis on the time cost}

% As shown in TABLE \ref{tab:TTA}, using AL on only $1 \%$ pixels of only $10 \%$ images in a batch, ODES outperforms SOTA TTA methods. Furthermore, as shown in Table \ref{tab:exp_decay}, exponentially decaying K with respect to the number of batches encountered, the model reaches peak performance after 60$\%$ batches are encountered, as such showing the scope of early stopping AL.

\section{\textbf{Conclusion}}

In this paper, we introduce ODES, a novel framework for online domain adaptation in medical image segmentation that uses active learning on a streaming data. In order to reduce the annotation burden of the active learner, ODES utilizes a unique image-pruning strategy which not only mitigates the challenge of domain shift, but also makes the application more online-friendly. This method ensures that only the most informative images are selected for annotation and so significantly reducing the time and effort required from experts. Through extensive experimentation, ODES has shown superior performance over existing TTA methods and also reaches close to the performance of offline adaptation. The conducted ablation analysis demonstrates the robustness and efficacy of our approach. 

\bibliographystyle{IEEEtran}
\bibliography{IEEEabrv,Bibliography}

\begin{thebibliography}{10}
\providecommand{\url}[1]{#1}
\csname url@samestyle\endcsname
\providecommand{\newblock}{\relax}
\providecommand{\bibinfo}[2]{#2}
\providecommand{\BIBentrySTDinterwordspacing}{\spaceskip=0pt\relax}
\providecommand{\BIBentryALTinterwordstretchfactor}{4}
\providecommand{\BIBentryALTinterwordspacing}{\spaceskip=\fontdimen2\font plus
\BIBentryALTinterwordstretchfactor\fontdimen3\font minus \fontdimen4\font\relax}
\providecommand{\BIBforeignlanguage}[2]{{%
\expandafter\ifx\csname l@#1\endcsname\relax
\typeout{** WARNING: IEEEtran.bst: No hyphenation pattern has been}%
\typeout{** loaded for the language `#1'. Using the pattern for}%
\typeout{** the default language instead.}%
\else
\language=\csname l@#1\endcsname
\fi
#2}}
\providecommand{\BIBdecl}{\relax}
\BIBdecl

\bibitem{liu2022region}
X.~Liu, L.~Yang, J.~Chen, S.~Yu, and K.~Li, ``Region-to-boundary deep learning model with multi-scale feature fusion for medical image segmentation,'' \emph{Biomedical Signal Processing and Control}, vol.~71, p. 103165, 2022.

\bibitem{campello2021multi}
V.~M. Campello, P.~Gkontra, C.~Izquierdo, C.~Martin-Isla, A.~Sojoudi, P.~M. Full, K.~Maier-Hein, Y.~Zhang, Z.~He, J.~Ma \emph{et~al.}, ``Multi-centre, multi-vendor and multi-disease cardiac segmentation: the m\&ms challenge,'' \emph{IEEE Transactions on Medical Imaging}, vol.~40, no.~12, pp. 3543--3554, 2021.

\bibitem{sharma2010automated}
N.~Sharma and L.~M. Aggarwal, ``Automated medical image segmentation techniques,'' \emph{Journal of medical physics/Association of Medical Physicists of India}, vol.~35, no.~1, p.~3, 2010.

\bibitem{lin2013online}
J.~F.-S. Lin and D.~Kuli{\'c}, ``Online segmentation of human motion for automated rehabilitation exercise analysis,'' \emph{IEEE Transactions on Neural Systems and Rehabilitation Engineering}, vol.~22, no.~1, pp. 168--180, 2013.

\bibitem{asgari2021deep}
S.~Asgari~Taghanaki, K.~Abhishek, J.~P. Cohen, J.~Cohen-Adad, and G.~Hamarneh, ``Deep semantic segmentation of natural and medical images: a review,'' \emph{Artificial Intelligence Review}, vol.~54, pp. 137--178, 2021.

\bibitem{ronneberger2015u}
O.~Ronneberger, P.~Fischer, and T.~Brox, ``U-net: Convolutional networks for biomedical image segmentation,'' in \emph{Medical Image Computing and Computer-Assisted Intervention--MICCAI 2015: 18th International Conference, Munich, Germany, October 5-9, 2015, Proceedings, Part III 18}.\hskip 1em plus 0.5em minus 0.4em\relax Springer, 2015, pp. 234--241.

\bibitem{anisha2015pragmatic}
P.~Anisha, C.~K.~K. Reddy, and L.~N. Prasad, ``A pragmatic approach for detecting liver cancer using image processing and data mining techniques,'' in \emph{2015 International Conference on Signal Processing and Communication Engineering Systems}.\hskip 1em plus 0.5em minus 0.4em\relax IEEE, 2015, pp. 352--357.

\bibitem{ibtehaz2020multiresunet}
N.~Ibtehaz and M.~S. Rahman, ``Multiresunet: Rethinking the u-net architecture for multimodal biomedical image segmentation,'' \emph{Neural networks}, vol. 121, pp. 74--87, 2020.

\bibitem{full2021studying}
P.~M. Full, F.~Isensee, P.~F. J{\"a}ger, and K.~Maier-Hein, ``Studying robustness of semantic segmentation under domain shift in cardiac mri,'' in \emph{Statistical Atlases and Computational Models of the Heart. M\&Ms and EMIDEC Challenges: 11th International Workshop, STACOM 2020, Held in Conjunction with MICCAI 2020}.\hskip 1em plus 0.5em minus 0.4em\relax Springer, 2021, pp. 238--249.

\bibitem{yang2022source}
C.~Yang, X.~Guo, Z.~Chen, and Y.~Yuan, ``Source free domain adaptation for medical image segmentation with fourier style mining,'' \emph{Medical Image Analysis}, vol.~79, p. 102457, 2022.

\bibitem{LE-UDA}
Z.~Zhao, F.~Zhou, K.~Xu, Z.~Zeng, C.~Guan, and S.~K. Zhou, ``Le-uda: Label-efficient unsupervised domain adaptation for medical image segmentation,'' \emph{IEEE Transactions on Medical Imaging}, vol.~42, no.~3, pp. 633--646, 2023.

\bibitem{li2021generalized}
S.~Li, B.~Xie, Q.~Lin, C.~H. Liu, G.~Huang, and G.~Wang, ``Generalized domain conditioned adaptation network,'' \emph{IEEE Transactions on Pattern Analysis and Machine Intelligence}, vol.~44, no.~8, pp. 4093--4109, 2021.

\bibitem{wei2021crest}
C.~Wei, K.~Sohn, C.~Mellina, A.~Yuille, and F.~Yang, ``Crest: A class-rebalancing self-training framework for imbalanced semi-supervised learning,'' in \emph{Proceedings of the IEEE/CVF conference on computer vision and pattern recognition}, 2021, pp. 10\,857--10\,866.

\bibitem{wang2022continual}
Q.~Wang, O.~Fink, L.~Van~Gool, and D.~Dai, ``Continual test-time domain adaptation,'' in \emph{Proceedings of the IEEE/CVF Conference on Computer Vision and Pattern Recognition}, 2022, pp. 7201--7211.

\bibitem{shin2021labor}
I.~Shin, D.-J. Kim, J.~W. Cho, S.~Woo, K.~Park, and I.~S. Kweon, ``Labor: Labeling only if required for domain adaptive semantic segmentation,'' in \emph{Proceedings of the IEEE/CVF International Conference on Computer Vision}, 2021, pp. 8588--8598.

\bibitem{xie2022towards}
B.~Xie, L.~Yuan, S.~Li, C.~H. Liu, and X.~Cheng, ``Towards fewer annotations: Active learning via region impurity and prediction uncertainty for domain adaptive semantic segmentation,'' in \emph{Proceedings of the IEEE/CVF Conference on Computer Vision and Pattern Recognition}, 2022, pp. 8068--8078.

\bibitem{he2021autoencoder}
Y.~He, A.~Carass, L.~Zuo, B.~E. Dewey, and J.~L. Prince, ``Autoencoder based self-supervised test-time adaptation for medical image analysis,'' \emph{Medical image analysis}, vol.~72, p. 102136, 2021.

\bibitem{hu2021fully}
M.~Hu, T.~Song, Y.~Gu, X.~Luo, J.~Chen, Y.~Chen, Y.~Zhang, and S.~Zhang, ``Fully test-time adaptation for image segmentation,'' in \emph{Medical Image Computing and Computer Assisted Intervention--MICCAI 2021: 24th International Conference, Strasbourg, France, September 27--October 1, 2021, Proceedings, Part III 24}.\hskip 1em plus 0.5em minus 0.4em\relax Springer, 2021, pp. 251--260.

\bibitem{wang2020tent}
D.~Wang, E.~Shelhamer, S.~Liu, B.~Olshausen, and T.~Darrell, ``Tent: Fully test-time adaptation by entropy minimization,'' \emph{arXiv preprint arXiv:2006.10726}, 2020.

\bibitem{niu2022efficient}
S.~Niu, J.~Wu, Y.~Zhang, Y.~Chen, S.~Zheng, P.~Zhao, and M.~Tan, ``Efficient test-time model adaptation without forgetting,'' in \emph{International conference on machine learning}.\hskip 1em plus 0.5em minus 0.4em\relax PMLR, 2022, pp. 16\,888--16\,905.

\bibitem{BatesonTTA}
M.~Bateson, H.~Lombaert, and I.~Ben~Ayed, ``Test-time adaptation with shape moments for image segmentation,'' in \emph{Medical Image Computing and Computer Assisted Intervention -- MICCAI 2022}.\hskip 1em plus 0.5em minus 0.4em\relax Cham: Springer Nature Switzerland, 2022, pp. 736--745.

\bibitem{karani2021test}
N.~Karani, E.~Erdil, K.~Chaitanya, and E.~Konukoglu, ``Test-time adaptable neural networks for robust medical image segmentation,'' \emph{Medical Image Analysis}, vol.~68, p. 101907, 2021.

\bibitem{zhang2023satta}
Y.~Zhang, K.~Huang, C.~Chen, Q.~Chen, and P.-A. Heng, ``Satta: Semantic-aware test-time adaptation for cross-domain medical image segmentation,'' in \emph{International Conference on Medical Image Computing and Computer-Assisted Intervention}.\hskip 1em plus 0.5em minus 0.4em\relax Springer, 2023, pp. 148--158.

\bibitem{li2023scale}
Z.~Li, J.~Yang, Y.~Xu, L.~Zhang, W.~Dong, and B.~Du, ``Scale-aware test-time click adaptation for pulmonary nodule and mass segmentation,'' in \emph{International Conference on Medical Image Computing and Computer-Assisted Intervention}.\hskip 1em plus 0.5em minus 0.4em\relax Springer, 2023, pp. 681--691.

\bibitem{yang2022dltta}
H.~Yang, C.~Chen, M.~Jiang, Q.~Liu, J.~Cao, P.~A. Heng, and Q.~Dou, ``Dltta: Dynamic learning rate for test-time adaptation on cross-domain medical images,'' \emph{IEEE Transactions on Medical Imaging}, vol.~41, no.~12, pp. 3575--3586, 2022.

\bibitem{hu2019overcoming}
W.~Hu, Z.~Lin, B.~Liu, C.~Tao, Z.~T. Tao, D.~Zhao, J.~Ma, and R.~Yan, ``Overcoming catastrophic forgetting for continual learning via model adaptation,'' in \emph{International conference on learning representations}, 2019.

\bibitem{kirkpatrick2017overcoming}
J.~Kirkpatrick, R.~Pascanu, N.~Rabinowitz, J.~Veness, G.~Desjardins, A.~A. Rusu, K.~Milan, J.~Quan, T.~Ramalho, A.~Grabska-Barwinska \emph{et~al.}, ``Overcoming catastrophic forgetting in neural networks,'' \emph{Proceedings of the national academy of sciences}, vol. 114, no.~13, pp. 3521--3526, 2017.

\bibitem{AL_ent2}
V.~Nath, D.~Yang, B.~A. Landman, D.~Xu, and H.~R. Roth, ``Diminishing uncertainty within the training pool: Active learning for medical image segmentation,'' \emph{IEEE Transactions on Medical Imaging}, vol.~40, no.~10, pp. 2534--2547, 2021.

\bibitem{wang2016cost}
K.~Wang, D.~Zhang, Y.~Li, R.~Zhang, and L.~Lin, ``Cost-effective active learning for deep image classification,'' \emph{IEEE Transactions on Circuits and Systems for Video Technology}, vol.~27, no.~12, pp. 2591--2600, 2016.

\bibitem{shen2018deep}
Y.~Shen, H.~Yun, Z.~C. Lipton, Y.~Kronrod, and A.~Anandkumar, ``Deep active learning for named entity recognition,'' in \emph{International Conference on Learning Representations}, 2018.

\bibitem{nath2020diminishing}
V.~Nath, D.~Yang, B.~A. Landman, D.~Xu, and H.~R. Roth, ``Diminishing uncertainty within the training pool: Active learning for medical image segmentation,'' \emph{IEEE Transactions on Medical Imaging}, vol.~40, no.~10, pp. 2534--2547, 2020.

\bibitem{joshi2009multi}
A.~J. Joshi, F.~Porikli, and N.~Papanikolopoulos, ``Multi-class active learning for image classification,'' in \emph{2009 ieee conference on computer vision and pattern recognition}.\hskip 1em plus 0.5em minus 0.4em\relax IEEE, 2009, pp. 2372--2379.

\bibitem{pmlr-v70-gal17a}
Y.~Gal, R.~Islam, and Z.~Ghahramani, ``Deep {B}ayesian active learning with image data,'' in \emph{Proceedings of the 34th International Conference on Machine Learning}, 2017, pp. 1183--1192.

\bibitem{gaillochet2023active}
M.~Gaillochet, C.~Desrosiers, and H.~Lombaert, ``Active learning for medical image segmentation with stochastic batches,'' \emph{Medical Image Analysis}, vol.~90, p. 102958, 2023.

\bibitem{xie2022active}
B.~Xie, L.~Yuan, S.~Li, C.~H. Liu, X.~Cheng, and G.~Wang, ``Active learning for domain adaptation: An energy-based approach,'' in \emph{Proceedings of the AAAI Conference on Artificial Intelligence}, vol.~36, no.~8, 2022, pp. 8708--8716.

\bibitem{hershey2007approximating}
J.~R. Hershey and P.~A. Olsen, ``Approximating the kullback leibler divergence between gaussian mixture models,'' in \emph{2007 IEEE International Conference on Acoustics, Speech and Signal Processing-ICASSP'07}, vol.~4.\hskip 1em plus 0.5em minus 0.4em\relax IEEE, 2007, pp. IV--317.

\bibitem{kavur2021chaos}
A.~E. Kavur, N.~S. Gezer, M.~Bar{\i}{\c{s}}, S.~Aslan, P.-H. Conze, V.~Groza, D.~D. Pham, S.~Chatterjee, P.~Ernst, S.~{\"O}zkan \emph{et~al.}, ``Chaos challenge-combined (ct-mr) healthy abdominal organ segmentation,'' \emph{Medical Image Analysis}, vol.~69, p. 101950, 2021.

\bibitem{macdonald_2022_6328447}
\BIBentryALTinterwordspacing
J.~A. Macdonald, Z.~Zhu, B.~Konkel, M.~Mazurowski, W.~Wiggins, and M.~Bashir, ``Duke liver dataset (mri),'' Mar. 2022. [Online]. Available: \url{https://doi.org/10.5281/zenodo.6328447}
\BIBentrySTDinterwordspacing

\bibitem{liu2020ms}
Q.~Liu, Q.~Dou, L.~Yu, and P.~A. Heng, ``Ms-net: multi-site network for improving prostate segmentation with heterogeneous mri data,'' \emph{IEEE transactions on medical imaging}, vol.~39, no.~9, pp. 2713--2724, 2020.

\bibitem{ramalho2012phase}
M.~Ramalho, V.~Her{\'e}dia, R.~O. de~Campos, B.~M. Dale, R.~M. Azevedo, and R.~C. Semelka, ``In-phase and out-of-phase gradient-echo imaging in abdominal studies: intra-individual comparison of three different techniques,'' \emph{Acta Radiologica}, vol.~53, no.~4, pp. 441--449, 2012.

\bibitem{chen2017deeplab}
L.-C. Chen, G.~Papandreou, I.~Kokkinos, K.~Murphy, and A.~L. Yuille, ``Deeplab: Semantic image segmentation with deep convolutional nets, atrous convolution, and fully connected {CRFs},'' \emph{IEEE Trans. Pattern Analysis and Machine Intelligence (TPAMI)}, 2017.

\bibitem{kingma2014adam}
D.~P. Kingma and J.~Ba, ``Adam: A method for stochastic optimization,'' \emph{arXiv preprint arXiv:1412.6980}, 2014.

\bibitem{shen2017deep}
Y.~Shen, H.~Yun, Z.~C. Lipton, Y.~Kronrod, and A.~Anandkumar, ``Deep active learning for named entity recognition,'' \emph{ICLR}, 2018.

\bibitem{culotta2005reducing}
A.~Culotta and A.~McCallum, ``Reducing labeling effort for structured prediction tasks,'' in \emph{AAAI}, vol.~5, 2005, pp. 746--751.

\end{thebibliography}

\end{document}